\pdfoutput=1
\documentclass{esannV2}

\usepackage{graphicx}
\usepackage[latin1]{inputenc}
\usepackage{amssymb,amsmath,array}
\usepackage{cite}
\usepackage{gensymb}
\usepackage{caption}
\usepackage{float}

\begin{document}
\title{Battery detection of XRay images using transfer learning}
\author{Nermeen Abou Baker$^1$ 	
David Rohrschneider$^1$	and Uwe Handmann$^1$
\thanks{This work has been funded by the Ministry of Economy, Innovation, Digitization, and Energy of the State of North Rhine-Westphalia within the project Prosperkolleg.}
\vspace{.3cm}\\
1- Ruhr West University of Applied Sciences - Dept of Computer Science \\
Lutzowstrassse $5$, $46236$ Bottrop - Germany\\
}
\maketitle
\begin{abstract}
The need for detecting and sorting batteries is drastically increasing for many applications. This study proves the potential of transfer learning in predicting whether the image contains a battery or not, the location and identifying three types of batteries, namely: prismatic, pouch, and cylindrical Lithium-Ion Batteries~(LIB). Particularly, it focuses on the transfer learning method in two applications: Training a large-scale dataset to detect electronic devices using a pre-trained YOLOv$5$m, then using these latter trained weights to detect and classify the batteries. The precision of battery detection achieves $94\%$, which outperforms the pre-trained YOLOv$5$m weights with $5\%$, in $22$ ms inference time.
\end{abstract}
\section{Introduction}
Inner feature screening is commonly performed in food processing, medical images, and airport baggage inspection for prohibited products using XRay images. The major benefit of using this technique is that any damage, dirt, or corrosion on the surface of the object does not affect the detection prediction of the internal components.\\
Battery recycling became a global problem with the increasing consumption of electronic devices~\cite{S-Cube-BakSzaHan2020}. Proper collection, storage, and sorting of waste can reduce hazardous substances entering the soil, water, and human body as well as reducing the cost of recycling process. Besides, these batteries contain expensive metals like silver, zinc, and cadmium, which are rare Earth elements. Manual battery inspection and shredding could cause fire or lead-acid batteries that could leak when damaged~\cite{NABTLSC}. Therefore, there is a need to tackle this problem using a smart and efficient strategy.\\
The rapid development in deep learning has enormously enhanced the ability to detect objects on images. Moreover, the ability of state-of-the-art deep learning structures boosts sorting accuracy, which surpasses human sorting efficiency. A study by~\cite{STERKENS2021105246}, found that identifying and sorting various types of batteries using XRay technology can be solved through deep learning. Each battery type has unique regions that are automatically detected in XRay. The study identifies the battery type despite the condition of the device, whether there is no label or marking on the battery, or whether the battery is rusted. Their system can predict whether the device contains a battery or not, the location, and six types of battery technologies, namely: cylindrical nickel-metal hydride, cylindrical alkaline, cylindrical zinc-carbon, cylindrical LIB, pouch LIB, and button cell batteries, with an $89\%$ precision and an $81\%$ recall of battery detection using YOLOv$2$ using their small-scale dataset.
\section{Implementation}
\begin{figure}[h!]
\centering
\includegraphics[width=0.95\textwidth,height=0.28\textheight]{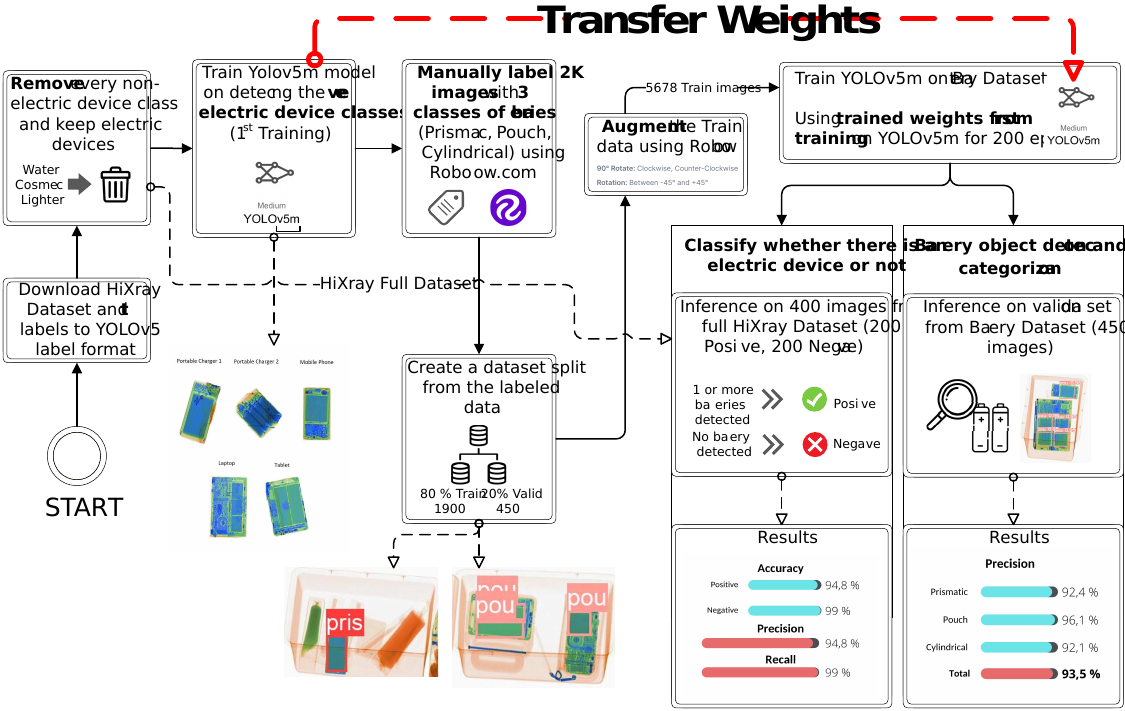}
\caption{Graphical abstract of the used method, implementation and results.}\label{VisAbs}
\end{figure}

Two experiments were conducted to detect whether there is an electrical device or not, the location of the battery and classifying the battery type, as shown in figure~\ref{VisAbs}.
The description of the used datasets and training the model using transfer learning will be described in this section.
\subsection{Datasets}
\subsubsection{Finding an appropriate dataset}
The first step in preparing the experiment was to find an XRay images dataset with electronic devices that contains multiple types of batteries.
There are three related datasets having at least one class of electronic devices: Durham DBF$6$~\cite{Gaus2019OnTU}, PIDRay~\cite{9710407}, and HiXray~\cite{Tao2021TowardsRX}. Since DBF$6$ dataset was not published as open-source and PIDRay contains one class of electronic devices (power bank), the HiXray dataset was the most suitable choice for our study. HiXray contains more than $45000$ high-quality XRay images with $102.928$ labels of $8$ different object classes from an international airport.
\subsubsection{Preparing the datasets for detecting the device and battery types}
\begin{figure}[H]
\centering
\includegraphics[width=0.8\textwidth, height=0.22\textheight]{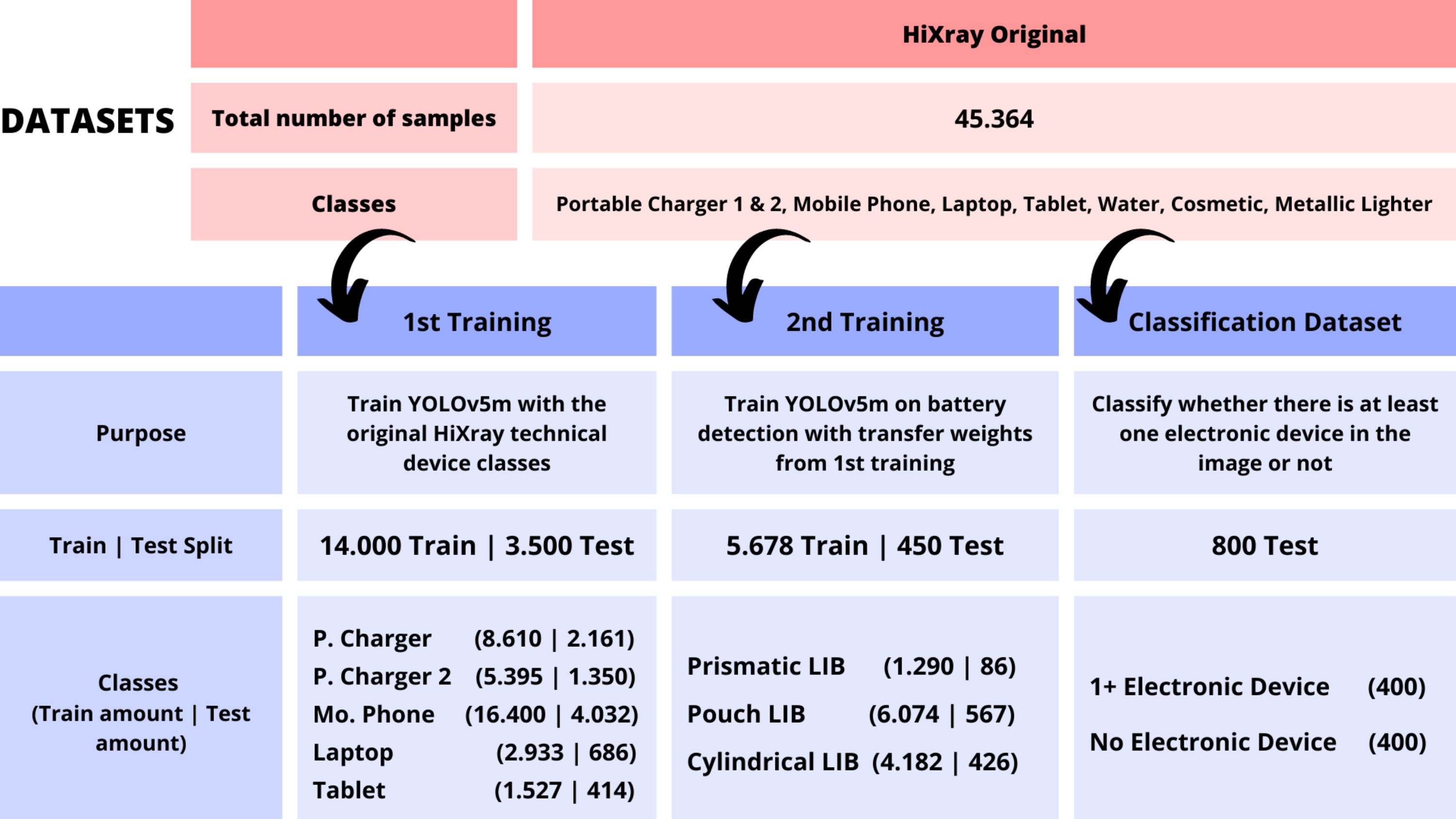}
\caption{The prepared datasets for this study.}\label{datasets}
\end{figure}

Three datasets were prepared, as shown in figure~\ref{datasets}, and they are described as follows:
\begin{itemize}
  \item For the first experiment: To use the HiXray data, each label has been converted to the YOLOv$5$ text-format. Then, all the irrelevant labels were removed, which are water, cosmetic and non-metallic lighters. So, a total of $17500$ samples were used, containing $15000$ samples with at least one class and $2000$ samples without electronic device, with split ratio $80\%$ for training and $20\%$ for testing.
  \item Creating the classification dataset for the purpose of sorting whether there is an electronic device or not. The set comprises 800 images of which $50\%$ contain at least one electronic device class and $50\%$ none.
  \item Creating the battery dataset with manual annotation: The resulting battery dataset contains $1900$ training samples and $450$ testing samples. For both, training and testing sets, there were $75\%$ samples with at least one and $25\%$ without any electronic device. RoboFlow~\cite{roboflow} was used in this study for manual annotation of batteries by drawing bounding boxes around each one. As a result, $5175$ batteries were labeled in a total of $2250$ images. The ratios of battery classes existing in the samples were $10\%$, $50\%$, $40\%$ for prismatic, pouch and cylindrical LIB cells, respectively. Data augmentation was applied as rotation $r \in \{$-45\degree$,~$+45\degree$,~$-90\degree$,~$+90\degree$\}$ to show the model different variations of batteries that could be found in the real test.
\end{itemize}
\subsection{Transfer learning applications}
For every training, YOLOv$5$m was used~\cite{jocher_2020}. It is the state-of-the-art regarding real-time deep learning for object detection. It is the latest and the middle-weight version of YOLO algorithms and uses PyTorch framework rather than DarkNet. The main improvement to the family of YOLO models is the Focus layer that replaces the first three layers of YOLOv$3$ to reduce the required CUDA memory and increase the forward propagation and backpropagation,with pre-trained weights on the COCO dataset~\cite{s22020464}. Battery detection is required for many real-time applications, like airport inspections, E-Waste recycling, etc. YOLO outperforms two-stage architecture like Faster R-CNN~\cite{NIPS2015_14bfa6bb},  Single Shot multi-box Detector (SSD)~\cite{SSD}, and RetinaNet \cite{Retin} in terms of speed and ability to perform real-time tasks accurately~\cite{survey19}.\\
The following hyper-parameters were used throughout this study: Learning rate $0.01$, image size $640$x$640$, with augmentations of: translation $0.1$, scaling $0.5$, flip $0.5$, and mosaic $1.0$.
For implementation, Google Colab was used with an NVIDIA Tesla P$100$-PCIE GPU.\\
In this study, transfer learning was applied in two phases:
\begin{itemize}
  \item Transfer learning using YOLOv$5$ weights: For the first training, the first dataset was used to detect the electronic devices classes on the images, based on the HiXray classes. The first experiment was performed using the weights of pre-trained YOLOv$5$m on COCO dataset, on $20$ Epochs with $32$ batch-size.
  \item Transfer learning using the weights from the first training in:
  \begin{itemize}
  \item Detecting the position and the battery class.
    \item Predicting the presence of the battery in the third dataset. The model was trained for $200$ Epochs with an early stop to prevent overfitting. When at least one battery has been detected from each image, the prediction was set to positive, otherwise, it was set to negative.
  \end{itemize}
\end{itemize}
\section{Results and discussion}
\begin{figure}[h!]
     \centering
\includegraphics[width=0.95\textwidth, height=0.2\textheight]{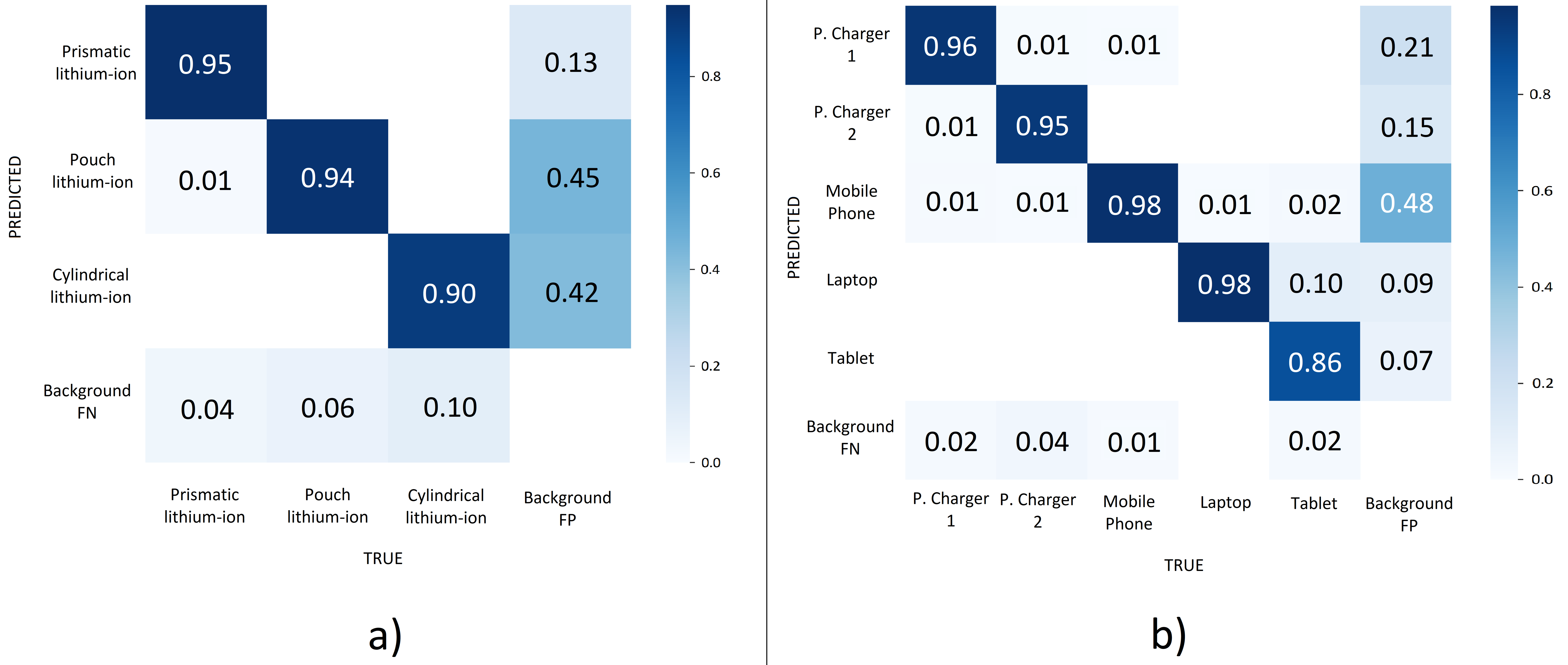}
\caption{Confusion matrices of object detection: (a) for batteries \& (b) for electrical devices.}\label{CM12}
\end{figure}

For the first experiment of training YOLOv$5$m on the first dataset, performed on electrical devices only, the results are shown in table \ref{TabExp1}, and figure \ref{CM12} (b).
\begin{table}[b!]
  \centering
  \small
  \begin{tabular}{|c|c|c|c|c|c|c|}
    \hline
      & PO$1$ &PO$2$  &Mobile phone &Laptop &Tablet &Total\\
    \hline
    Training& $8610$ & $5395$&\emph{16400}&$2933$&$1527$&$34865$\\
    Validation& $2161$ & $1350$& \emph{4032}  &$686$& $414$& $8643$\\
    Precision on Val Data&\textbf{0.966} & $0.96$ & $0.939$ & $0.899$ & $0.936$  &\textbf{0.94} \\
    Recall on Val Data& $0.945$ &$0.921$ &$0.965$ & \textbf{0.99} &$0.867$ &$0.939$ \\
    F1-Score &\textbf{0.955} & $0.94$ & $0.952$ & $0.945$ & $0.9$ & $0.938$\\ \hline
  \end{tabular}
  \caption{Evaluation of electrical device detection.}\label{TabExp1}
\end{table}
For the second experiment of using the previously trained weights (transfer learning), tested for battery detection. The combination of the results of table \ref{TabExp2}, and figure \ref{CM12} (a), shows, that the transferred weights achieves an overall precision $94\%$, which outperforms the YOLOv$5$m weights. However, the cylindrical LIB has a relatively lower performance because they have small size, which could be confused with some noisy background objects. In addition, by using these weights, the classification of electronic and non-electronic devices are shown in table \ref{classification}. The trained model for the second experiment took $22$ ms for each inference, which is great for real-time testing.\\
\begin{table}[h!]
  \centering
  \small
  \begin{tabular}{|c|c|c|c|c|}
    \hline
   &   Prismatic LIB &Pouch LIB  &Cylindrical LIB & Total\\
      \hline
Training &$1290$ &\emph{6074} &$4182$ &$11546$\\
Validation&$86$  &\emph{567} & $426$ &$1079$ \\\hline
 &\multicolumn{4}{c|} {With YOLOv5m weights}\\ \hline
Precision on Val Data& $0.818$ & {0.962} & $0.899$ &\textbf{0.893} \\
Recall on Val Data& {0.812} &$0.901$ & $0.772$ & $0.828$ \\
F1-Score & $0.815$ & {0.931} &$0.831$ & $0.859$ \\ \hline
 &\multicolumn{4}{c|} {With the transferred weights}\\ \hline
Precision on Val Data & $0.924$ & \textbf{0.961} & $0.921$ &\textbf{0.935} \\
Recall on Val Data & \textbf{0.948} &$0.929$ & $0.892$ & $0.923$ \\
F1-Score & $0.936$ & \textbf{0.945} &$0.906$ & $0.929$ \\ \hline
  \end{tabular}
  \caption{Evaluation of battery detection using the YOLOv$5$ weights and our transferred trained weights.}\label{TabExp2}
\end{table}
\begin{table}[h!]
  \centering
  \small
  \begin{tabular}{|c|c|c|}
    \hline
     & Electronic Device &  No Electronic Device \\
    \hline
    Accuracy on Validation Data & $0.9475$ &$0.99$\\
    Precision on Validation Data & \multicolumn{2}{c|} {\textbf{0.948}} \\
    Recall on Validation Data & \multicolumn{2}{c|} {$0.99$} \\
    F$1$-Score on Validation Data & \multicolumn{2}{c|} {$0.969$}\\
    \hline
  \end{tabular}
  \caption{Evaluation classifying the devices into electrical and non-electrical.}\label{classification}
\end{table}
The original HiXray dataset suffers from unbalancing where the dominance of the mobile phone class exists in most of the images. Moreover, mobile phones, tablets, and most of the laptops in the images contain multiple pouch LIB cells. Therefore, it was not possible to achieve a numerical balance between the three battery classes. Furthermore, the occlusion of objects makes it difficult to manually label the batteries in some images and results in lower detection performance. Despite these challenges, previous results show that detecting electronic devices achieves $94\%$ overall precision. Furthermore, the transferred weights outperform the YOLOv$5$m trained weights by $5\%$ in detecting batteries and predicting the presence of electrical devices with $95\%$ precision. 
\section{Conclusion}
This study shows that transfer learning is an efficient method for detecting electrical devices and batteries in XRay images. Transfer learning was used in two experiments: Detecting electronic devices using the weights of the pre-trained YOLOv$5$m model with a total precision of $94\%$, then transferring these trained weights to detect and classify the battery with a precision of $94\%$, which shows that it outperforms the results of using the YOLOv$5$m weights with a precision of $89\%$, running $22$ms for each inference. Moreover, the study achieves $95\%$ of predicting the presence of electrical devices.
\begin{footnotesize}
\bibliographystyle{unsrt}
\bibliography{ref}

@article{STERKENS2021105246,
title = {Detection and recognition of batteries on X-Ray images of waste electrical and electronic equipment using deep learning},
journal = {Resources, Conservation and Recycling},
volume = {168},
pages = {105246},
year = {2021},
issn = {0921-3449},
doi = {https://doi.org/10.1016/j.resconrec.2020.105246},
url = {https://www.sciencedirect.com/science/article/pii/S0921344920305619},
author = {Wouter Sterkens and Dillam Diaz-Romero and Toon Goedeme and Wim Dewulf and Jef R. Peeters}
}

@Article{s22020464,
AUTHOR = {Nepal, Upesh and Eslamiat, Hossein},
TITLE = {Comparing YOLOv3, YOLOv4 and YOLOv5 for Autonomous Landing Spot Detection in Faulty UAVs},
JOURNAL = {Sensors},
VOLUME = {22},
YEAR = {2022},
NUMBER = {2},
ARTICLE-NUMBER = {464},
URL = {https://www.mdpi.com/1424-8220/22/2/464},
PubMedID = {35062425},
ISSN = {1424-8220},
DOI = {10.3390/s22020464}
}

@article{survey19,
  author    = {Licheng Jiao and
               Fan Zhang and
               Fang Liu and
               Shuyuan Yang and
               Lingling Li and
               Zhixi Feng and
               Rong Qu},
  title     = {A Survey of Deep Learning-based Object Detection},
  journal   = {CoRR},
  volume    = {abs/1907.09408},
  year      = {2019},
  url       = {http://arxiv.org/abs/1907.09408},
  eprinttype = {arXiv},
  eprint    = {1907.09408},
  bibsource = {dblp computer science bibliography, https://dblp.org}
}

@ARTICLE{NABTLSC,
    author={Nermeen Abou Baker and Paul Szabo-M\'{y}ller and Uwe Handmann},
    title={Transfer learning-based method for automated e-waste recycling in smart cities},
    journal={EAI Endorsed Transactions on Smart Cities},
    volume={5},
    number={16},
    publisher={EAI},
    journal_a={SC},
    year={2021},
    month={4},
    doi={10.4108/eai.16-4-2021.169337}
}

@conference{S-Cube-BakSzaHan2020,
  author = {Nermeen Abou Baker and Paul Szabo-M\'{y}ller and Uwe Handmann},
  booktitle = {EAI S-CUBE 2020 - 11th EAI International Conference on Sensor Systems and Software},
  title = {{Feature-fusion transfer learning method as a basis to support automated smartphone recycling in a circular smart city}},
  year = {2020},
  address = {online},
  url = {https://builder.eai.eu/share/52890}
}

@article{Gaus2019OnTU,
  title={On the Use of Deep Learning for the Detection of Firearms in X-ray Baggage Security Imagery},
  author={Yona Falinie Abdul Gaus and Neelanjan Bhowmik and T. Breckon},
  journal={2019 IEEE International Symposium on Technologies for Homeland Security (HST)},
  year={2019},
  pages={1-7}
}

@INPROCEEDINGS {9710407,
author = {Boying Wang and
               Libo Zhang and
               Longyin Wen and
               Xianglong Liu and
               Yanjun Wu},
booktitle = {2021 IEEE/CVF International Conference on Computer Vision (ICCV)},
title = {Towards Real-World Prohibited Item Detection: A Large-Scale X-ray Benchmark},
year = {2021},
volume = {},
issn = {},
pages = {5392-5401},
doi = {10.1109/ICCV48922.2021.00536},
url = {https://doi.ieeecomputersociety.org/10.1109/ICCV48922.2021.00536},
publisher = {IEEE Computer Society},
address = {Los Alamitos, CA, USA},
month = {oct}
}

@article{Tao2021TowardsRX,
  title={Towards Real-world X-ray Security Inspection: A High-Quality Benchmark And Lateral Inhibition Module For Prohibited Items Detection},
  author={Renshuai Tao and Yanlu Wei and Xiangjian Jiang and Hainan Li and Haotong Qin and Jiakai Wang and Yuqing Ma and Libo Zhang and Xianglong Liu},
  journal={2021 IEEE/CVF International Conference on Computer Vision (ICCV)},
  year={2021},
  pages={10903-10912}
}

@inproceedings{NIPS2015_14bfa6bb,
 author = {Ren, Shaoqing and He, Kaiming and Girshick, Ross and Sun, Jian},
 booktitle = {Advances in Neural Information Processing Systems},
 editor = {C. Cortes and N. Lawrence and D. Lee and M. Sugiyama and R. Garnett},
 pages = {},
 publisher = {Curran Associates, Inc.},
 title = {Faster R-CNN: Towards Real-Time Object Detection with Region Proposal Networks},
 url = {https://proceedings.neurips.cc/paper/2015/file/14bfa6bb14875e45bba028a21ed38046-Paper.pdf},
 volume = {28},
 year = {2015}
}

@InProceedings{SSD,
author="Liu, Wei
and Anguelov, Dragomir
and Erhan, Dumitru
and Szegedy, Christian
and Reed, Scott
and Fu, Cheng-Yang
and Berg, Alexander C.",
editor="Leibe, Bastian
and Matas, Jiri
and Sebe, Nicu
and Welling, Max",
title="SSD: Single Shot MultiBox Detector",
booktitle="Computer Vision -- ECCV 2016",
year="2016",
publisher="Springer International Publishing",
address="Cham",
pages="21--37",
isbn="978-3-319-46448-0"
}

@ARTICLE {Retin,
author = {Tsung{-}Yi Lin and
               Priya Goyal and
               Ross B. Girshick and
               Kaiming He and
               Piotr Doll{\'{a}}r},
journal = {IEEE Transactions on Pattern Analysis $\&$ Machine Intelligence},
title = {Focal Loss for Dense Object Detection},
year = {2020},
volume = {42},
number = {02},
issn = {1939-3539},
pages = {318-327},
doi = {10.1109/TPAMI.2018.2858826},
publisher = {IEEE Computer Society},
address = {Los Alamitos, CA, USA},
month = {feb}
}

@misc{jocher_2020, title={ultralytics/yolov5}, url={https://github.com/ultralytics/yolov5}, journal={GitHub}, author={Jocher, Glenn}, year={2020}, month={Jun}}

@misc{roboflow, title={https://roboflow.com/}, url={https://roboflow.com/}, journal={Roboflow}}
\end{footnotesize}
\end{document}